\documentclass[a4paper]{article}

\usepackage{INTERSPEECH2019}
\usepackage{url}
\usepackage{xcolor}
\usepackage[normalem]{ulem}

\usepackage{bbm}
\usepackage{subcaption}
\usepackage{tikz}
\usepackage{tikz-cd}
\usetikzlibrary{calc}
\usetikzlibrary{arrows}
\usetikzlibrary{shapes}
\usetikzlibrary{positioning}
\usetikzlibrary{decorations.pathmorphing}
\tikzstyle{every picture}+=[remember picture]
\usepackage{pgfplots}

\makeatletter
\renewcommand{\paragraph}{%
  \@startsection{paragraph}{4}%
  {\z@}{1.75ex \@plus 1ex \@minus .2ex}{-1em}%
  {\normalfont\normalsize\bfseries}%
}
\makeatother

\title{Towards Transfer Learning for End-to-End Speech Synthesis \\ from Deep Pre-Trained Language Models}
\name{Wei Fang, Yu-An Chung, James Glass}
\address{
  MIT Computer Science and Artificial Intelligence Laboratory\\
  Cambridge, MA 02139, USA}
\email{\{weifang,andyyuan,glass\}@mit.edu}

\pgfplotsset{compat=1.5}
\begin{document}

\maketitle

\begin{abstract}
Modern text-to-speech~(TTS) systems are able to generate audio that sounds almost as natural as human speech.
However, the bar of developing high-quality TTS systems remains high since a sizable set of studio-quality~$<$text, audio$>$ pairs is usually required.
Compared to commercial data used to develop state-of-the-art systems, publicly available data are usually worse in terms of both quality and size.
Audio generated by TTS systems trained on publicly available data tends to not only sound less natural, but also exhibits more background noise.
In this work, we aim to lower TTS systems' reliance on high-quality data by providing them the textual knowledge extracted by deep pre-trained language models during training.
In particular, we investigate the use of BERT to assist the training of Tacotron-2, a state of the art TTS consisting of an encoder and an attention-based decoder.
BERT representations learned from large amounts of unlabeled text data are shown to contain very rich semantic and syntactic information about the input text, and have potential to be leveraged by a TTS system to compensate the lack of high-quality data.
We incorporate BERT as a parallel branch to the Tacotron-2 encoder with its own attention head.
For an input text, it is simultaneously passed into BERT and the Tacotron-2 encoder.
The representations extracted by the two branches are concatenated and then fed to the decoder.
As a preliminary study, although we have not found incorporating BERT into Tacotron-2 generates more natural or cleaner speech at a human-perceivable level, we observe improvements in other aspects such as the model is being significantly better at knowing when to stop decoding such that there is much less babbling at the end of the synthesized audio and faster convergence during training.
\end{abstract}

\vspace{1em}

\section{Introduction}
\par
End-to-end, deep learning-based approaches are causing a paradigm shift in the field of text-to-speech~(TTS)~\cite{wang2017tacotron,arik2017deep,sotelo2017char2wav,shen2018natural,ping2019clarinet}.
Unlike traditional parametric TTS systems~\cite{zen2009statistical,taylor2009text}, which typically pipeline a text front-end, a duration model, an acoustic feature prediction model, and a vocoder, end-to-end neural TTS approaches integrate all these components into a single network and thus only require paired text and audio for training.
This removes the need of extensive domain expertise for designing each component individually, offering a much simplified voice building pipeline.
More importantly, they have been shown to be capable of generating speech that sounds almost as natural as humans~\cite{shen2018natural,li2019neural}.

\par
TTS systems that produce speech close to human quality, however, are usually trained on data collected by individual institutions that are not publicly accessible.
Due to their strictly-controlled conditions during the collection process, it is a common belief that these internal data have higher quality than the publicly available ones.
Such a belief is supported by the fact that when training modern TTS models such as Tacotron-2~\cite{shen2018natural} on publicly available dataset like LJ Speech~\cite{ito17ljspeech}, the synthesized speech sounds less natural and exhibits more noise in the audio background than that produced by Tacotron-2 trained on Google's internal data.
Therefore, it is still unclear whether we can achieve state-of-the-art TTS performance using only publicly available data.

\par
Recently, deep pre-trained language models~(LMs)~\cite{peters2018deep,howard2018universal,radford2018improving,devlin2019bert} have shown to be capable of extracting textual representations that contain very rich syntactic and semantic information about the text sequences.
By transferring the textual knowledge contained in these deep pre-trained LMs~(e.g., by replacing the original text input with the extracted text features), a simple downstream model is able to achieve state-of-the-art performance on a wide range of natural language processing tasks such as natural language inference, sentiment analysis, and question answering, to name a few.
These deep LMs are first trained on large amounts of unlabeled text data using self-supervised objectives, and then fine-tuned with task-specific losses along with models for the downstream tasks.

\par
In this work, we aim to leverage the textual knowledge contained in these deep pre-trained LMs to lower end-to-end TTS systems' reliance on high-quality data.
Particularly, we investigate the use of BERT~\cite{devlin2019bert} for assisting the training of Tacotron-2~\cite{shen2018natural}.
The backbone of Tacotron-2 is a sequence-to-sequence network~\cite{sutskever2014sequence} with attention~\cite{chorowski2015attention} that consists of an encoder and a decoder.
The goal of the encoder is to transform the input text into robust sequential representations of text, which are then consumed by the attention-based decoder for predicting the spectral features.
We enrich the textual information to be consumed by the decoder by feeding the linguistic features extracted by BERT from the same input text to the decoder as well along with the original encoder representations.

\par
Existing work such as~\cite{chung2019semi,ming2019feature} also attempts to make use of the textual knowledge learned from large text corpora to improve TTS.
However, they either focus on improving the data efficiency of TTS training, i.e., to minimize the amount of paired text and audio for training, in a small-data regime, or the word embedding modules they use to provide textual knowledge have some limitations by nature, e.g., the word embeddings are trained based on very shallow language modeling tasks and are hence considered less powerful than the one we use in this work.





\tikzstyle{block} = [rectangle, minimum width=0.1cm, minimum height=0.1cm, text centered, draw=black, rounded corners=0.07cm]
\tikzstyle{desc} = [minimum width=0.1cm, minimum height=0.1cm, text centered]
\tikzstyle{thinarrow} = [->]
\tikzstyle{thinline} = [-,>=stealth]

\begin{figure*}[t]
\centering
  \begin{tikzpicture}[]
    \footnotesize
    \node (charemb) [block,minimum width=3.5cm,minimum height=0.7cm] {Character Embedding};
    \node (text) [desc,below=0.8cm of charemb] {Input text};
    \node (conv) [block,minimum width=3.5cm,minimum height=0.7cm,above=0.25cm of charemb] {Convolution Layers};
    \node (enclstm) [block,minimum width=3.5cm,minimum height=0.7cm,above=0.25cm of conv] {Bi-directional LSTM};
    \node (att1) [block,minimum width=2.4cm,minimum height=1.5cm,right=1cm of conv,text width=2.35cm] {Location-sensitive Attention};

    \node (declstm) [block,minimum width=3.5cm,minimum height=0.7cm,right=4.5cm of conv] {Autoregressive LSTM};
    \node (prenet) [block,minimum width=3.5cm,minimum height=0.7cm,below=0.25cm of declstm] {Pre-Net};
    \node (postnet) [block,minimum width=2.45cm,minimum height=0.7cm,above=0.25cm of declstm.north west,anchor=south west] {Post-Net};
    \node (spec) [desc,minimum width=2.45cm,minimum height=0.7cm,above=1.6cm of postnet] {Mel Spectrogram};
    \node (stop) [desc,minimum width=0.95cm,minimum height=0.7cm,above=2.55cm of declstm.north east,anchor=south east] {Stop};
    \node (wavenet) [block,minimum width=2.45cm,minimum height=0.7cm,above=0.65cm of spec.north west,anchor=south west] {WaveGlow Vocoder};
    \node (wav) [desc,above=0.25cm of wavenet] {Waveform};

    \node (wpemb) [block,minimum width=3.5cm,minimum height=0.7cm,above=1.65cm of enclstm] {WordPiece Embedding};
    \node (trans) [block,minimum width=3.5cm,minimum height=0.7cm,above=0.25cm of wpemb] {Transformer layers};
    \node (att2) [block,minimum width=2.4cm,minimum height=1.5cm,right=1cm of wpemb.north east,text width=2.35cm] {Location-sensitive Attention};

    \draw [thinarrow] (text.north) to (charemb.south);
    \draw [thinarrow] (charemb.north) to (conv.south);
    \draw [thinarrow] (conv.north) to (enclstm.south);
    \draw [thinarrow] (enclstm.east) to [out=0,in=180] (att1.west);
    \draw [thinarrow] (att1.east) to [out=0,in=180] (declstm.west);
    \draw [thinarrow] (prenet.north) to (declstm.south);
    \draw [thinarrow] ($(postnet.south)-(0,0.25)$) to (postnet.south);
    \draw [thinarrow] (declstm.east) to [out=0,in=0] (prenet.east);
    \draw [thinarrow] (postnet.north) to (spec.south);
    \draw [thinarrow] ($(stop.south)-(0,2.55)$) to (stop.south);
    \draw [thinarrow] (spec.north) to [out=90,in=-90] (wavenet.south);
    \draw [thinarrow] (wavenet.north) to (wav.south);

    \draw [thinarrow] (text.north) -- ++(0,0.3) -- ++(-2.6,0) -- ++(0,4.5) -- ++(2.6,0) -- (wpemb.south);
    \draw [thinarrow] (wpemb.north) to (trans.south);
    \draw [thinarrow] (trans.east) to [out=0,in=180] (att2.west);
    \draw [thinarrow] (att2.east) to [out=-50,in=150] (declstm.west);

    \begin{scope}
      \draw [rounded corners=0.01cm, draw=black!65,dashed] ($(charemb.south west)+(-0.6,-0.35)$) rectangle ($(declstm.north east)+(0.6,2.1)$) {};
      \draw [rounded corners=0.01cm, draw=black!65,dashed] ($(wpemb.south west)+(-0.6,-0.2)$) rectangle ($(trans.north east)+(0.6,0.6)$) {};
      \draw [rounded corners=0.01cm, draw=black!125,dotted] ($(charemb.south west)+(-0.3,-0.16)$) rectangle ($(enclstm.north east)+(0.3,0.51)$) {};
      \draw [rounded corners=0.01cm, draw=black!125,dotted] ($(prenet.south west)+(-3.8,-0.16)$) rectangle ($(postnet.north east)+(1.45,0.51)$) {};
      \node (taco) [desc,above=0.55cm of enclstm.north west,anchor=south west] {\textbf{Original Tacotron2}};
      \node (bert) [desc,above=0.15cm of trans.north west,anchor=south west] {\textbf{BERT Encoder}};
      
      \node (enc) [desc,above=0.08cm of enclstm.north west,anchor=south west] {\textit{Encoder}};
      \node (dec) [desc,above=1.38cm of att1.west,anchor=south west] {\textit{Decoder}};
    \end{scope}
  \end{tikzpicture}
  \caption{Overall architecture of the proposed model. The bottom block depicts the original Tacotron-2, which is used as our base TTS system. Tacotron-2 consists of an encoder and an attention-based decoder. For an input text, the goal of the encoder is to transform it into robust textual representations that are then consumed by the attention-based decoder for predicting the spectral features. The upper part of the figure illustrates how we incorporate BERT into the TTS system. Specifically, BERT is fed with the same input text and extracts another sequence of textual representations of it. For each decoding time step, the decoder attends back to the two sequences of textual representations produced by the two branches~(i.e., the Tacotron-2 encoder and BERT) with separate attention heads. The two generated attention context vectors are concatenated and fed to the decoder as the new input for decoding.}
  \label{fig:arch}
\end{figure*}
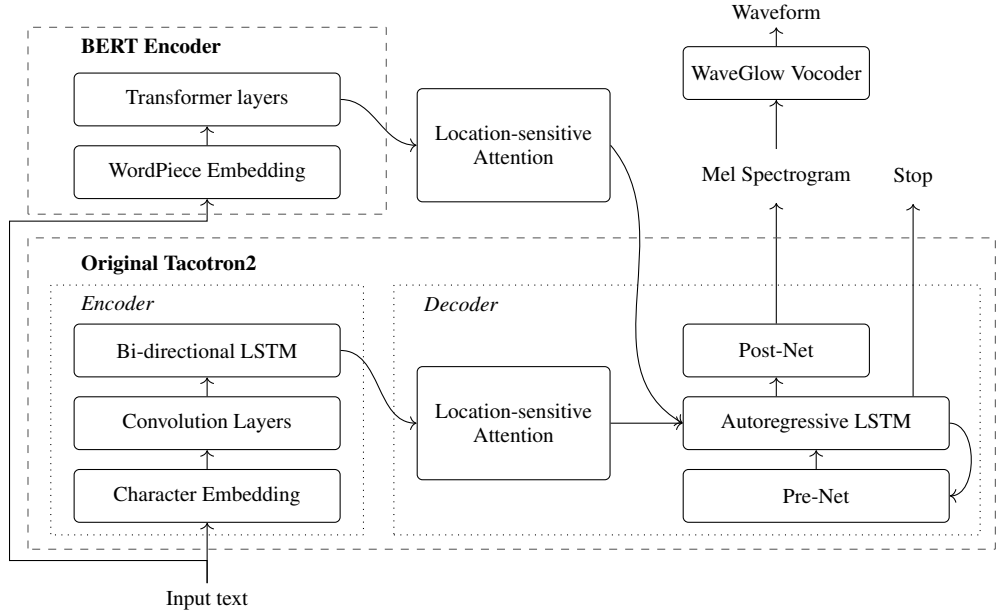

\section{Proposed Approach}
In this section, we start with introducing BERT~\cite{devlin2019bert} and Tacotron-2~\cite{shen2018natural}.
We then present the proposed approach for incorporating BERT representations into the training of Tacotron-2.
The proposed approach is illustrated in Figure~\ref{fig:arch}.

\subsection{Tacotron-2}
Tacotron-2 follows the sequence-to-sequence~(seq2seq) with attention framework and functions as a spectral feature~(e.g., mel spectrogram) prediction network.
The predicted spectral features are then inverted by a separately trained vocoder into time-domain waveforms.
In this work, we modify the seq2seq component by incorporating BERT as a parallel branch to the Tacotron-2 encoder, and use WaveGlow~\cite{prenger2019waveglow} as the vocoder to synthesize the audio waveforms.

\par
The seq2seq component consists of two main building blocks: an encoder and an attention-based decoder.
At a high level, the encoder takes a text sequence as input and transforms it into a sequence of textual representations.
For each decoding time step, the decoder conditions on these textual representations and predicts a mel spectrogram frame.

\par
Specifically, the encoder takes a character sequence as input, passes it through a stack of convolutional layers followed by a single-layer bidirectional LSTM, and the hidden states of the recurrent network are used as the encoded representations.
These representations will then be consumed by the decoder. 
The decoder is an autoregressive LSTM with the location-sensitive attention~\cite{chorowski2015attention} mechanism that summarizes the encoded representations at each decoding time step. 
The input to the decoder at each time step, which is the prediction from the previous step, is first passed through a {\it pre-net} before being fed into the LSTM. 
Lastly, the output of the LSTM is processed by a convolutional {\it post-net} to predict the final spectrogram.
The output of the LSTM is also fed into another network with a sigmoid activation to determine when to stop decoding.

\par
During training, the mean squared error is calculated at the spectrogram prediction output, while the stop-token prediction is trained with binary cross-entropy loss.
Additionally, teacher-forcing is used to train the recurrent decoder.
During inference, the ground truth targets are unknown. 
The decoder generates the mel spectrogram in an autoregressive fashion as opposed to teacher-forcing.
Generation is completed when the stop token output exceeds a threshold of~$0.5$.

\subsection{BERT}
\par
Our goal is to leverage rich textual knowledge contained in deep pre-trained LMs to assist TTS training.
To do so, we use BERT to transform the input text sequence into textual representations that are in parallel to those extracted by the Tacotron-2 encoder, and provide both of them to the Tacotron-2 decoder.
BERT is a Transformer-based~\cite{vaswani2017attention} model trained on large amounts of text in an unsupervised manner.
Below we briefly summarize BERT.
\paragraph*{Input Representations.}
Unlike Tacotron-2 that takes character sequence as input, the input to BERT is a sequence of subword units that are usually referred to as WordPieces~\cite{wu2016google}, where the tokenization is determined by a Byte-Pair Encoding process~\cite{sennrich2016neural}.
Since a Transformer is position-agnostic, it needs positional information encoded in their input, thus it uses learnable positional embeddings for WordPiece tokens with length up to~$512$.
Additionally, a special {\tt [CLS]} token is added at the beginning of each sequence so that the final hidden state corresponding to this token represents the aggregate vector of the sequence.
\paragraph*{The Transformer Model.}
The backbone of BERT is a multi-layer bidirectional Transformer encoder, which consists of multiple blocks of multi-head self-attention stacked together.
In this work we use the $\mathrm{BERT}_\mathrm{BASE}$ configuration that contains~$12$ attention blocks each with~$12$ attention heads and hidden size of~$768$.
\paragraph*{Pre-Training.}
The BERT model is pre-trained using two unsupervised objectives: masked language modeling, for which the model has to predict a randomly masked out token, and next sentence prediction, where two sentences are packed as input to the encoder and the aggregate vector is used to predict whether they are adjacent sentences.
\subsection{Using BERT in Tacotron-2 Framework}
\par
As mentioned previously, the Tacotron-2 encoder aims to extract robust sequential representations of the input text, the decoder then decodes the spectrogram by conditioning on these textual representations.
By drawing analogy to the traditional parametric TTS systems, the Tacotron-2 encoder can be viewed as the linguistic front-end and the decoder is similar to the statistical acoustic prediction model.
\par
From this view, we propose to inject the textual information contained in the BERT representations to the Tacotron-2 decoder, so that it has access to the textual features from both the Tacotron-2 encoder and BERT to make a spectral prediction.
In practice, we feed the input text, which is first tokenized into WordPiece sequence, into BERT, and the representations from the last attention block is exposed to the decoder.
The decoder then attends back to both the Tacotron-2 encoder representations and BERT representations using separate location-sensitive attention heads in order to produce the corresponding attention context vectors.
The two vectors are then concatenated before feeding into the autoregressive recurrent network.

\par
There are other ways for incorporating external textual representations into the text front-end of a TTS system, for example by concatenating the representations with the character embedding sequence~\cite{chung2019semi}.
However, there usually exists a mismatch between the time resolutions of the two sequences and the existing solutions are not as intuitive as simply letting the decoder attend to the representations extracted by both branches.
%
%

\par
Note that although we use Tacotron-2 as the base TTS system in this work, our framework can be easily extended to other TTS systems as well.

\section{Experiments}
\subsection{Dataset and Settings}
\par
We use LJ Speech~\cite{ito17ljspeech}, a public domain speech dataset consisting of 13100 audio clips of a single speaker reading from non-fiction books.
The audio utterances vary in length from 1 to 10 seconds, with a total length of about 24 hours of speech.
They are encoded with 16-bit PCM with a sample rate of $22050$Hz.
The dataset is split into train, validation, and test sets of $12500$, $100$, and $500$ clips respectively.

\par
We use the implementation of Tacotron-2 by Nvidia\footnote{\url{https://github.com/NVIDIA/tacotron2}} and keep the default hyperparameters to train the baseline model.
For our model we use the same configuration for the Tacotron-2 component.
The parameterization of the additional location-sensitive attention layer for attending to the BERT representations is also kept the same.
The entire model is trained end-to-end, thus the losses are also back-propagated into the BERT encoder to fine-tune the textual representations.
The Adam optimizer is used with a learning rate of $0.001$ to learn the parameters.

\subsection{Evaluation Metrics}
\par
To measure performance, we use several metrics used in previous works that correlate with voice quality and prosody.
To compare the generated audio to the reference audio, these metrics are only calculated up to the length of the shorter audio signal.
The pitch and voicing metrics are computed using the YIN~\cite{de2002yin} pitch tracking algorithm.

\paragraph*{Mean Cepstral Distortion (MCD${}_K$)~\cite{kubichek1993mel}:}
$$\mathrm{MCD}_K=\frac{1}{T}\sum_{t=0}^{T-1}\sqrt{\sum_{k=1}^K(c_{t,k}-\hat{c}_{t,k})^2},$$
where $c_{t,k}$ and $\hat{c}_{t,k}$ are the $k$-th mel frequency cepstral coefficient (MFCC) of the $t$-th frame from the reference and generated audio, respectively. The overall energy $c_{t,0}$ is discarded in this measure.
We follow previous work and set $K=13$.

\paragraph*{Gross Pitch Error (GPE)~\cite{nakatani2008method}:}
$$\mathrm{GPE}=\frac{\sum_t \mathbbm{1}[|p_t-\hat{p}_t|>0.2 p_t]\mathbbm{1}[v_t]\mathbbm{1}[\hat{v}_t]}{\sum_t \mathbbm{1}[v_t]\mathbbm{1}[\hat{v}_t]},$$
where $p_t$ and $\hat{p}_t$ are the pitch signals from the reference and generated audio, $v_t$ and $\hat{v}_t$ are the voicing decisions from the reference and generated audio, and $\mathbbm{1}$ denotes the indicator function.
GPE measures the percentage of voiced frames that deviate by more than 20\% in the pitch signal of the generated audio compared to the reference.

\begin{table}[t]
    \caption{Performance of our model versus the baseline Tacotron-2 model on several evaluation metrics.}
    \centering
    \begin{tabular}{cccc}
      \hline
        Model & MCD${}_{13}$ & GPE & FFE \\
      \hline
        Tacotron-2 & $21.88$ & $0.722$ & $0.740$ \\
      \hline
        Ours & $25.21$ & $0.720$ & $0.735$ \\
      \hline
    \end{tabular}
    \label{tab:res}
\end{table}

\paragraph*{F0 Frame Error (FFE)~\cite{chu2009reducing}:}
$$\mathrm{FFE}=\frac{1}{T}\sum_t \mathbbm{1}[|p_t-\hat{p}_t|>0.2 p_t]\mathbbm{1}[v_t]\mathbbm{1}[\hat{v}_t]+\mathbbm{1}[v_t\neq \hat{v}_t].$$
Following the definitions of GPE, FFE measures the percentage of frames that either have a 20\% pitch error or a differing voicing decision between the generated and reference audio.

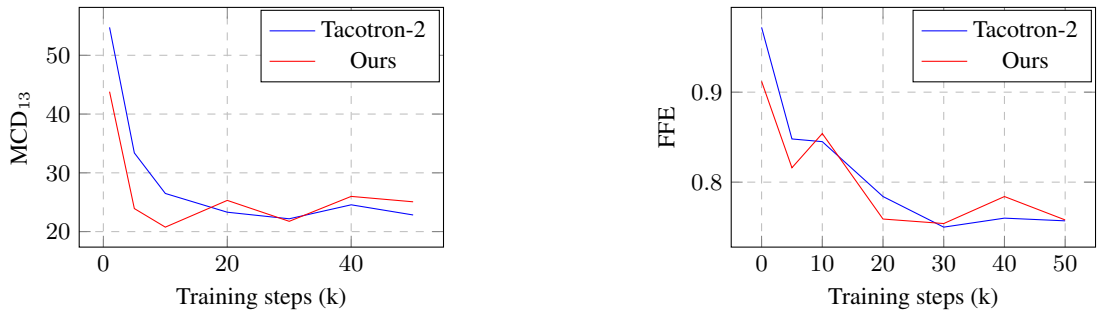
\begin{figure*}[htbp]
  \begin{minipage}{0.5\linewidth}
  \hspace{1.7em}
  \begin{tikzpicture}
    \begin{axis}[
      xlabel={Training steps (k)},
      ylabel={MCD${}_{13}$},
      xmajorgrids=true,
      ymajorgrids=true,
      xmajorgrids=true,
      grid style=dashed,
      width=.75\textwidth,
      height=.56\textwidth,
      legend style={at={(0.99,0.99)},anchor=north east}
    ]
    \addplot[
      color of colormap=1,
      ]
      coordinates {
        (1,54.75)(5,33.39)(10,26.49)
        (20,23.3)(30,22.18)(40,24.56)(50,22.83)
      };
    \addlegendentry{Tacotron-2}
    \addplot[
      color of colormap=1000,
      ]
      coordinates {
        (1,43.81)(5,23.92)(10,20.76)
        (20,25.31)(30,21.75)(40,25.99)(50,25.07)
      };
    \addlegendentry{Ours}
    \end{axis}
  \end{tikzpicture}
  \end{minipage}
  \begin{minipage}{.5\linewidth}
  \hspace{1.6em}
  \begin{tikzpicture}
    \begin{axis}[
      xtick={0,10,20,30,40,50},
      xlabel={Training steps (k)},
      ylabel={FFE},
      xmajorgrids=true,
      ymajorgrids=true,
      xmajorgrids=true,
      grid style=dashed,
      width=.75\textwidth,
      height=.56\textwidth,
      legend style={at={(0.99,0.99)},anchor=north east}
    ]
    \addplot[
      color of colormap=1,
      ]
      coordinates {
        (0,0.972)
        (5,0.848)(10,0.845)
        (20,0.784)(30,0.750)(40,0.760)(50,0.757)
      };
    \addlegendentry{Tacotron-2}
    \addplot[
      color of colormap=1000,
      ]
      coordinates {
        (0,0.912)
        (5,0.816)(10,0.854)
        (20,0.759)(30,0.754)(40,0.784)(50,0.758)
      };
    \addlegendentry{Ours}
    \end{axis}
  \end{tikzpicture}
  \end{minipage}
    \caption{Plots for ${MCD}_{13}$ and FFE metrics as training steps increases.}
    \label{fig:curve}
\end{figure*}

\begin{figure*}[htbp]
    \centering
    \begin{subfigure}[t]{.49\linewidth}
        \centering
        \includegraphics[width=.8\linewidth]{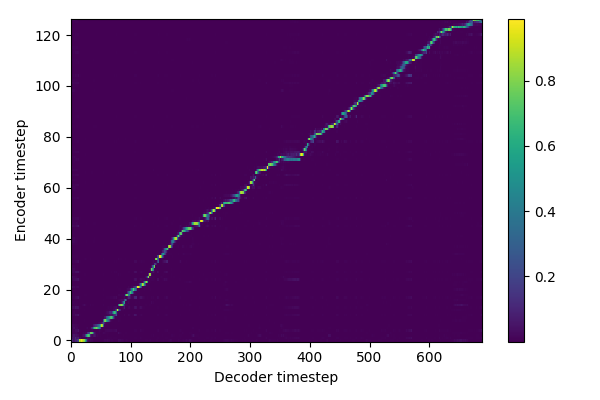}
        \caption{Encoder Attention}
        \label{fig:attenc}
    \end{subfigure}
    \hfill
    \begin{subfigure}[t]{.49\linewidth}
        \centering
        \includegraphics[width=.8\linewidth]{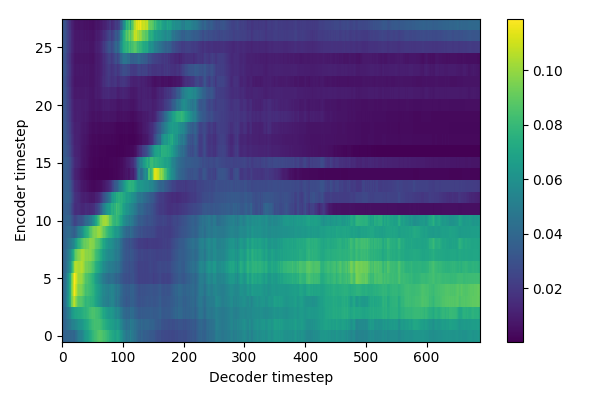}
        \caption{BERT Attention}
        \label{fig:attbert}
    \end{subfigure}
    \caption{Attention alignments for encoder and BERT encoder on a test phrase.}
    \label{fig:attention}
\end{figure*}

\subsection{Results and Observations}
Table \ref{tab:res} shows the performance of our model versus the baseline Tacotron-2 model without the BERT encoder.
All metrics calculate some form of error of the generated audio against the reference audio, thus the lower the better.
On the MCD${}_{13}$ metric, our model performs a little worse than the baseline model, but on the GPE and FFE metrics our model performs slightly better.
Overall, there is not much difference in terms of these metrics at the end of training between both models.
Additionally, we do not find a difference in naturalness for audios synthesized by the two models at the end.

\par
However, if we visualize the performance metrics against the number of training steps, as shown in Figure \ref{fig:curve}, we see that our model converges faster compared to the baseline Tacotron-2.
This suggests that the addition of the learned representations of BERT may have helped our model learn quicker early on during training.
Even though the metrics converge to similar values at the end of training, our model still benefits from the pre-trained knowledge from BERT.
We also discover a difference early on in training in terms of the quality of the audio generated by the two models.
The quality of the audio generated by our model is generally better than the baseline Tacotron-2, but the discrepancy disappears as training progresses.
These results are to some extent similar to those reported in previous work~\cite{chung2019semi}, where semi-supervised learning with external text representations and pre-training improves synthesis performance in low-resource settings but not in large-data regimes.

\par
From the learning curves, we can also see that the MCD${}_{13}$ metric converges much quicker than the FFE metric and oscillates within a small range, and we do not observe much correlation of the MCD${}_{13}$ metric with audio quality.
We even find that models at early training stages that achieve relatively low MCD${}_{13}$ synthesize gibberish audio.
Compared to MCD${}_{13}$, we find that FFE offers a much better indication of audio quality.

\par
Additionally, we observe a clear improvement in predicting when to stop generation with our model.
The baseline Tacotron-2 often has trouble ending its decoding process, generating gibberish noise at the end.
Our model, on the other hand, almost never exhibits this behavior.

\par
We visualize the attention alignments of the decoder against both attention layers of our model in Figure \ref{fig:attention}.
From Figure \ref{fig:attenc}, we see that the attention is a almost diagonal, meaning that the decoder mostly focuses on the correct characters as the audio sequence is generated.
This pattern is also observed in the original Tacotron~\cite{wang2017tacotron}, as well as the baseline Tacotron-2 model.
From what we observe, this behavior is strongly correlated with synthesized audio quality, but unfortunately there isn't a straightforward method to quantify this behavior.
On the other hand, in our model the attention layer that attends to the BERT representations, shown in Figure \ref{fig:attbert}, only has a rough diagonal pattern at the beginning of decoding.
Its attention is more spread across different time steps compared to the encoder attention alignments.
We can also see from the figures that the values of the attention are much lower for the BERT attention.
Since the encoder attention patterns are similar, we hypothesize that the model still learns to map the text sequence to acoustic feature sequence mainly by the textual representations learned from the encoder.
The BERT representations serve as additional information that the decoder uses to improve its prediction.

\section{Discussion and Future Work}
In this work, we propose to exploit the textual representations from pre-trained deep LMs for improving end-to-end neural speech synthesis.
As a preliminary study, we have not found incorporating BERT into the Tacotron-2 framework significantly improves the quality of the synthesized audio;
however, we do find that our approach improves the Tacotron-2 model in other aspects such as faster convergence during training and the final model is significantly better at knowing when to stop decoding such that the synthesized audio has less babbling in the end.

\par
This is only a preliminary work, and there is still a lot left to be studied.
%
%
For instance, instead of utilizing only knowledge from unlabeled text data, we can also make use of large amounts of unlabeled speech corpora~\cite{hsu2019disentangling} to assist the decoder in learning the acoustic representations and alignments.
A potential method could be to initialize Tacotron decoder with a pre-trained auto-regressive predictive coding model~\cite{chung2019unsupervised}.

\bibliographystyle{IEEEtran}
\bibliography{main}

\begin{thebibliography}{10}
\providecommand{\url}[1]{#1}
\csname url@samestyle\endcsname
\providecommand{\newblock}{\relax}
\providecommand{\bibinfo}[2]{#2}
\providecommand{\BIBentrySTDinterwordspacing}{\spaceskip=0pt\relax}
\providecommand{\BIBentryALTinterwordstretchfactor}{4}
\providecommand{\BIBentryALTinterwordspacing}{\spaceskip=\fontdimen2\font plus
\BIBentryALTinterwordstretchfactor\fontdimen3\font minus
  \fontdimen4\font\relax}
\providecommand{\BIBforeignlanguage}[2]{{%
\expandafter\ifx\csname l@#1\endcsname\relax
\typeout{** WARNING: IEEEtran.bst: No hyphenation pattern has been}%
\typeout{** loaded for the language `#1'. Using the pattern for}%
\typeout{** the default language instead.}%
\else
\language=\csname l@#1\endcsname
\fi
#2}}
\providecommand{\BIBdecl}{\relax}
\BIBdecl

\bibitem{wang2017tacotron}
Y.~Wang, R.~Skerry-Ryan, D.~Stanton, Y.~Wu, R.~Weiss, N.~Jaitly, Z.~Yang,
  Y.~Xiao, Z.~Chen, S.~Bengio, Q.~Le, Y.~Agiomyrgiannakis, R.~Clark, and
  R.~Saurous, ``Tacotron: Towards end-to-end speech synthesis,'' in
  \emph{Interspeech}, 2017.

\bibitem{arik2017deep}
S.~{\"O}. Arik, M.~Chrzanowski, A.~Coates, G.~Diamos, A.~Gibiansky, Y.~Kang,
  X.~Li, J.~Miller, A.~Ng, J.~Raiman, S.~Sengupta, and M.~Shoeybi, ``Deep
  voice: Real-time neural text-to-speech,'' in \emph{ICML}, 2017.

\bibitem{sotelo2017char2wav}
J.~Sotelo, S.~Mehri, K.~Kumar, J.~F. Santos, K.~Kastner, A.~Courville, and
  Y.~Bengio, ``Char2wav: End-to-end speech synthesis,'' in \emph{ICLR
  Workshop}, 2017.

\bibitem{shen2018natural}
J.~Shen, R.~Pang, R.~Weiss, M.~Schuster, N.~Jaitly, Z.~Yang, Z.~Chen, Y.~Zhang,
  Y.~Wang, R.~Skerrv-Ryan, R.~Saurous, Y.~Agiomyrgiannakis, and Y.~Wu,
  ``Natural {TTS} synthesis by conditioning wavenet on mel spectrogram
  predictions,'' in \emph{ICASSP}, 2018.

\bibitem{ping2019clarinet}
W.~Ping, K.~Peng, and J.~Chen, ``Clarinet: Parallel wave generation in
  end-to-end text-to-speech,'' in \emph{ICLR}, 2019.

\bibitem{zen2009statistical}
H.~Zen, K.~Tokuda, and A.~Black, ``Statistical parametric speech synthesis,''
  \emph{Speech Communication}, vol.~51, no.~11, pp. 1039--1064, 2009.

\bibitem{taylor2009text}
P.~Taylor, \emph{Text-to-speech synthesis}.\hskip 1em plus 0.5em minus
  0.4em\relax Cambridge University Press, 2009.

\bibitem{li2019neural}
N.~Li, S.~Liu, Y.~Liu, S.~Zhao, M.~Liu, and M.~Zhou, ``Neural speech synthesis
  with transformer network,'' in \emph{AAAI}, 2019.

\bibitem{ito17ljspeech}
K.~Ito, ``The {LJ} speech dataset,''
  \url{https://keithito.com/LJ-Speech-Dataset/}, 2017.

\bibitem{peters2018deep}
M.~Peters, M.~Neumann, M.~Iyyer, M.~Gardner, C.~Clark, K.~Lee, and
  L.~Zettlemoyer, ``Deep contextualized word representations,'' in
  \emph{NAACL-HLT}, 2018.

\bibitem{howard2018universal}
J.~Howard and S.~Ruder, ``Universal language model fine-tuning for text
  classification,'' in \emph{ACL}, 2018.

\bibitem{radford2018improving}
A.~Radford, K.~Narasimhan, T.~Salimans, and I.~Sutskever, ``Improving language
  understanding by generative pre-training,'' OpenAI, Tech. Rep., 2018.

\bibitem{devlin2019bert}
J.~Devlin, M.-W. Chang, K.~Lee, and K.~Toutanova, ``{BERT}: Pre-training of
  deep bidirectional transformers for language understanding,'' in
  \emph{NAACL-HLT}, 2019.

\bibitem{sutskever2014sequence}
I.~Sutskever, O.~Vinyals, and Q.~Le, ``Sequence to sequence learning with
  neural networks,'' in \emph{NIPS}, 2014.

\bibitem{chorowski2015attention}
J.~Chorowski, D.~Bahdanau, D.~Serdyuk, K.~Cho, and Y.~Bengio, ``Attention-based
  models for speech recognition,'' in \emph{NIPS}, 2015.

\bibitem{chung2019semi}
Y.-A. Chung, Y.~Wang, W.-N. Hsu, Y.~Zhang, and R.~Skerry-Ryan,
  ``Semi-supervised training for improving data efficiency in end-to-end speech
  synthesis,'' in \emph{ICASSP}, 2019.

\bibitem{ming2019feature}
H.~Ming, L.~He, H.~Guo, and F.~Soong, ``Feature reinforcement with word
  embedding and parsing information in neural {TTS},'' \emph{arXiv preprint
  arXiv:1901.00707}, 2019.

\bibitem{prenger2019waveglow}
R.~Prenger, R.~Valle, and B.~Catanzaro, ``Waveglow: A flow-based generative
  network for speech synthesis,'' in \emph{ICASSP}, 2019.

\bibitem{vaswani2017attention}
A.~Vaswani, N.~Shazeer, N.~Parmar, J.~Uszkoreit, L.~Jones, A.~Gomez,
  {\L}.~Kaiser, and I.~Polosukhin, ``Attention is all you need,'' in
  \emph{NIPS}, 2017.

\bibitem{wu2016google}
Y.~Wu, M.~Schuster, Z.~Chen, Q.~Le, M.~Norouzi, W.~Macherey, M.~Krikun, Y.~Cao,
  Q.~Gao, K.~Macherey, J.~Klingner, A.~Shah, M.~Johnson, X.~Liu, Å.~Kaiser,
  S.~Gouws, Y.~Kato, T.~Kudo, H.~Kazawa, K.~Stevens, G.~Kurian, N.~Patil,
  W.~Wang, C.~Young, J.~Smith, J.~Riesa, A.~Rudnick, O.~Vinyals, G.~Corrado,
  M.~Hughes, and J.~Dean, ``Google's neural machine translation system:
  Bridging the gap between human and machine translation,'' \emph{arXiv
  preprint arXiv:1609.08144}, 2016.

\bibitem{sennrich2016neural}
R.~Sennrich, B.~Haddow, and A.~Birch, ``Neural machine translation of rare
  words with subword units,'' in \emph{ACL}, 2016.

\bibitem{de2002yin}
A.~De~Cheveign{\'e} and H.~Kawahara, ``{YIN}, a fundamental frequency estimator
  for speech and music,'' \emph{The Journal of the Acoustical Society of
  America}, vol. 111, no.~4, pp. 1917--1930, 2002.

\bibitem{kubichek1993mel}
R.~Kubichek, ``Mel-cepstral distance measure for objective speech quality
  assessment,'' in \emph{PacRim}, 1993.

\bibitem{nakatani2008method}
T.~Nakatani, S.~Amano, T.~Irino, K.~Ishizuka, and T.~Kondo, ``A method for
  fundamental frequency estimation and voicing decision: Application to infant
  utterances recorded in real acoustical environments,'' \emph{Speech
  Communication}, vol.~50, no.~3, pp. 203--214, 2008.

\bibitem{chu2009reducing}
W.~Chu and A.~Alwan, ``Reducing f0 frame error of f0 tracking algorithms under
  noisy conditions with an unvoiced/voiced classification frontend,'' in
  \emph{ICASSP}, 2009.

\bibitem{hsu2019disentangling}
W.-N. Hsu, Y.~Zhang, R.~Weiss, Y.-A. Chung, Y.~Wang, Y.~Wu, and J.~Glass,
  ``Disentangling correlated speaker and noise for speech synthesis via data
  augmentation and adversarial factorization,'' in \emph{ICASSP}, 2019.

\bibitem{chung2019unsupervised}
Y.-A. Chung, W.-N. Hsu, H.~Tang, and J.~Glass, ``An unsupervised autoregressive
  model for speech representation learning,'' \emph{arXiv preprint
  arXiv:1904.03240}, 2019.

\end{thebibliography}

\end{document}